\documentclass[11pt]{article}

\usepackage[margin=1in]{geometry}
\usepackage{amsmath}
\usepackage{amssymb}
\usepackage{amsthm}
\usepackage{mathtools}
\usepackage{bm}

\usepackage{graphicx}
\usepackage{booktabs}
\usepackage{multirow}
\usepackage{subcaption}
\usepackage{tikz}
\usetikzlibrary{arrows.meta, positioning, fit, backgrounds, shapes.geometric}
\usepackage{float}
\usepackage{algorithm}
\usepackage{algpseudocode}

\PassOptionsToPackage{colorlinks=true,
            linkcolor=blue!60!black,
            citecolor=green!50!black,
            urlcolor=blue!60!black}{hyperref}
\usepackage{orcidlink}

\usepackage{microtype}
\usepackage{xcolor}
\usepackage{url}
\usepackage{xspace}

\usepackage[numbers,sort&compress]{natbib}
\usepackage{hyperref}

\theoremstyle{definition}

\newtheorem{principle}{Principle}
\theoremstyle{remark}

\newcommand{\cwl}{\textsc{CWL}\xspace}

\title{\textbf{Beyond Compaction: Structured Context Eviction for Long-Horizon Agents}}

\author{
  Andrew Semenov\orcidlink{0009-0009-7047-5179}\thanks{Correspondence to \texttt{andysem3@gmail.com} (permanent personal address).} \\
  Kiz8 \\
  \texttt{ands@kiz8.team} $\cdot$ \texttt{andysem3@gmail.com}
  \and
  Svyatoslav Dorofeev\orcidlink{0009-0004-3480-1743} \\
  Kiz8 \\
  \texttt{sean@kiz8.team}
}

\date{April 21, 2026}

\begin{document}
\maketitle

\begin{abstract}
We present Context Window Lifecycle (\cwl), a context-management scheme that gives long-horizon LLM agents an effectively unbounded working horizon. As a session accumulates history, \cwl\ keeps the context within budget through graduated, semantically-aware eviction: the agent annotates its trajectory as typed, dependency-linked \emph{episodes} as work proceeds, and a deterministic, LLM-free policy evicts content in priority order within that structure when a token budget is exceeded. \cwl\ preserves user turns and the exploratory context the agent is actively reasoning over, while aggressively shedding action episodes whose effects are already persisted in the environment, keeping active context near a stable ceiling — itself below the regime where attention degrades and hallucination rates rise.

Compared to summarization-based compaction, \cwl\ avoids four well-known limitations: unpredictable lossiness, destruction of causal structure, blocking model cost, and compression-induced hallucination. Compared to recency truncation, \cwl\ is semantically aware: it drops the oldest-and-most-recoverable content according to the dependency graph rather than oldest-in-time regardless of relevance. We describe the annotation protocol, the episode graph, the eviction policy, and the token-accounting loop — demonstrated empirically by a single agent session completing 89 sequential tasks across 80 million tokens with no measurable degradation in task accuracy relative to per-task isolated sessions.
\end{abstract}

\section{Introduction}
\label{sec:intro}

Long-running LLM agents --- coding assistants, research agents, tool-using workflows --- face a structural problem: the context window is finite, but the trajectory of work is not. Every tool call, every file read, every retrieved document accumulates in the history, and the model's effective reasoning budget shrinks with each turn. Without intervention, the session either terminates when the window fills or begins silently dropping the oldest content, taking with it whatever context was needed to make later decisions coherent.

The prevailing intervention is \emph{compaction}: when the history approaches the window limit, the agent is paused, the accumulated transcript is handed to an LLM with instructions to summarize it, and the summary replaces the original history. The agent then resumes against this compressed substrate. Compaction is attractive in its simplicity --- it requires no structural assumptions about the interaction --- and it is now the default in several widely used agent frameworks.

Summarization-based compaction has four well-known limitations, each of which compounds under the conditions in which it fires --- mid-task, under token pressure:

\begin{enumerate}
  \item \textbf{Lossiness is unpredictable.} The summary is produced by an LLM under its own constraints; what is retained and what is dropped depends on the summarizer's instantaneous judgment of salience, which need not align with what the downstream agent will require. Errors are not detectable from the compacted context alone.
  \item \textbf{Structure is destroyed.} The original trajectory contains explicit causal structure: a tool call produced an output, that output informed a decision, that decision produced an action. Prose summaries collapse this into narrative, erasing the provenance that would let the agent revisit its own reasoning.
  \item \textbf{Compression is expensive and blocking.} A compaction pass is a full LLM call over a large portion of the context window, adding latency measured in seconds and cost measured in tokens-in-tokens-out. It happens precisely when the agent is mid-task.
  \item \textbf{Hallucinations are introduced at the worst moment.} Summarization under length pressure is a known failure mode for LLMs. Compaction introduces novel errors into the context at the exact point where the agent has the least remaining budget to detect and correct them.
\end{enumerate}

\textbf{Context Window Lifecycle} (\cwl) addresses these limitations and delivers a capability neither prior approach achieves: a long-horizon agent that can operate indefinitely without coherence loss. Rather than treating the transcript as an opaque blob to be summarized when full or a queue to be truncated from the front, \cwl\ treats it as a \emph{structured record of work} that the agent has annotated as it went. When the budget is exceeded, a deterministic policy walks that structure and evicts content in priority order, starting with the most recoverable and ending, if necessary, with whole episodes.

The key insight is that the agent is in the best possible position to annotate its own trajectory --- it knows, at the moment of doing the work, which parts are the live context it still needs and which parts are action records whose effects are already written to the environment, and it knows which exploration a given action depended on. \cwl\ provides the agent with a single \texttt{delimiter} tool to express this structure incrementally, and the compression policy exploits it.

\paragraph{Contributions.}
\begin{itemize}
  \item We demonstrate that long-horizon LLM agents can operate with an effectively unbounded working horizon: a single session completing 89 sequential tasks across 80 million tokens with no measurable degradation in task accuracy relative to per-task isolated sessions.
  \item We propose \cwl, a context-management scheme built around three primitives: \emph{typed episodes}, an \emph{explicit dependency graph} authored by the agent, and a \emph{graduated eviction policy} that is deterministic and LLM-free.
  \item We characterize four well-known limitations of summarization-based compaction (unpredictable lossiness, structural destruction, blocking cost, and compression-induced hallucination) and show how \cwl's design addresses each by construction.
  \item We describe the full architecture: the annotation protocol exposed to the agent, the episode graph's invariants, the eviction priority ordering, and the token accounting loop that triggers eviction.
\end{itemize}

\paragraph{Scope of this paper.} This is the first release of \cwl. The design, architecture, and initial empirical results are presented here in full. Larger-scale evaluations --- additional benchmark suites, higher run counts, and extended ablations --- are planned for a follow-up version. We release now to establish the approach and make its design decisions legible for others to adopt, critique, or build on.

\section{Related Work}
\label{sec:related}

\paragraph{Summarization-based compaction.} The dominant approach in contemporary agent frameworks is to trigger a summarization pass when the context crosses a threshold (often 70--90\% of the window)~\citep{packer2023memgpt, kang2025acon, wu2025resum, lindenbauer2025complexity}. \cwl\ differs in that compression is incremental, deterministic, and does not invoke a model; the cost is paid in annotation discipline during normal operation rather than in a large, blocking summarization at the end.

\paragraph{Context-Folding.} Sun et al.~\citep{sun2025contextfolding} introduce two agent actions, \texttt{branch} and \texttt{return}, that impose a two-level plan-execute hierarchy on the trajectory. When the agent encounters a token-intensive subtask, it calls \texttt{branch(description, prompt)}, opening a clean sub-context; when the subtask is done, it calls \texttt{return(message)}, collapsing the intermediate steps and retaining only the model-written summary in the main thread. A further engineering consequence follows at the attention layer: \texttt{return} rolls back the KV-cache to the state prior to the \texttt{branch} call, so the sub-context is discarded both at the token level and in the inference engine's cached activations. The resulting compression is substantial --- the main trajectory is held near 8{,}000 tokens while the agent processes over 100{,}000 in total. Because a sparse outcome reward is insufficient for learning effective branch-and-return behaviour, the authors train their agent with FoldGRPO, a dense process-reward variant of GRPO that adds three token-level signals: a penalty when the main thread exceeds 50\% of the working limit without branching, a scope-deviation penalty assessed by a judge model when branch actions stray from the declared sub-task, and a failure penalty for turns with failed tool calls.

The structural intuition behind Context-Folding overlaps with \cwl: both approaches recognize that agentic work has a natural hierarchical structure --- exploration/investigation vs. committing to outcomes --- and exploit that structure to decide what to compress. The differences are consequential, however. First, \emph{who decides}: in Context-Folding the model learns when to compress and what to retain through RL training, while in \cwl\ the eviction policy is a deterministic algorithm consulting an explicit dependency graph --- no model is invoked. This distinction has two downstream effects: Context-Folding requires fine-tuning a specific model and is therefore not model-agnostic, and the preserved \texttt{return} message is model-generated content, which reintroduces the hallucination risk that \cwl\ avoids by construction. Second, \emph{depth}: Context-Folding enforces a strict two-level hierarchy (nesting is disabled inside an active branch), whereas \cwl's dependency graph can in principle represent arbitrary-depth causal chains. Third, \emph{trigger}: Context-Folding's compression is implicit --- the trained model is expected to branch proactively before the context fills --- while \cwl's trigger is explicit and reactive, firing deterministically from token accounting.

\paragraph{Recency truncation and sliding windows.} The simplest approach is to drop the oldest turns until the context fits~\citep{xiao2023streamingllm, lindenbauer2025complexity}. This is cheap and predictable but semantically blind: it will happily evict the tool call that defined a variable the agent is still using. \cwl\ can be viewed as a structurally-informed generalization: the ``oldest'' content is not the oldest in time but the oldest-and-most-recoverable in the dependency graph.

\paragraph{Retrieval-augmented context.} An alternative is to keep the full history in external storage and retrieve relevant passages at each turn~\citep{park2023generative}. This addresses a different problem from \cwl\ --- selective injection of external knowledge versus keeping a live session within budget --- and the two are orthogonal.

\paragraph{Agentic memory systems.} A broader literature explores persistent memory for LLM agents~\citep{packer2023memgpt, xu2025amem, chhikara2025mem0, rasmussen2025zep}, typically focused on cross-session state. \cwl\ addresses the complementary within-session problem of keeping a single trajectory coherent under a token budget.

\section{Design Principles}
\label{sec:principles}

Before describing the architecture, we state the principles that guided it. These are not claims the paper will prove; they are commitments the design is intended to honor, and every subsequent decision can be traced back to one of them.

\begin{principle}[Compression is part of the protocol, not a recovery action]
\label{prin:protocol}
The agent does not operate normally until the context fills and then invokes a repair. It operates with the understanding that compression is ongoing, annotates its trajectory to support it, and never encounters a state in which the context must be rescued.
\end{principle}

\begin{principle}[The agent is the authority on structure]
\label{prin:authority}
The system infers as little as possible about what the transcript means. Episode boundaries, episode types, and dependencies are declared explicitly by the agent using dedicated tools. 
\end{principle}

\begin{principle}[User content is inviolable]
\label{prin:user}
Content authored by the human participant is never evicted, regardless of token pressure. If the budget cannot be met without touching user turns, the system surfaces the condition rather than silently degrading.
\end{principle}

\begin{principle}[Causal dependencies dominate recency]
\label{prin:causal}
An old episode that a recent decision depended on is more valuable than a recent episode that stands alone. Eviction order follows the dependency graph, not the timeline.
\end{principle}

\begin{principle}[Compression must not invoke the model]
\label{prin:llm-free}
Every step of the eviction policy is deterministic and local. This rules out compression-induced hallucination by construction and keeps the cost of a compression pass at effectively zero.
\end{principle}

\begin{principle}[Graduated, not catastrophic]
\label{prin:graduated}
Compression proceeds in the smallest increments that will meet the budget. The first response to overflow is to strip the single most recoverable piece of content, not to restructure the session.
\end{principle}

\section{Architecture}
\label{sec:arch}

\subsection{Overview}

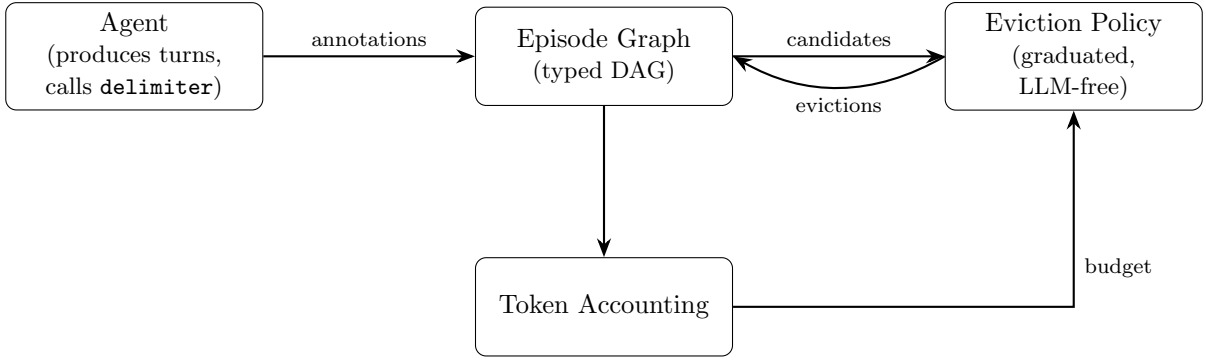
\begin{figure}[t]
  \centering
  \begin{tikzpicture}[
    box/.style={draw, rounded corners, align=center, minimum height=1.3cm, minimum width=3.4cm, font=\small},
    arr/.style={-{Stealth[length=2.5mm]}, thick}
  ]
    \node[box] (agent) {Agent \\ \footnotesize(produces turns,\\\footnotesize calls \texttt{delimiter})};
    \node[box, right=2.8cm of agent] (graph) {Episode Graph \\ \footnotesize(typed DAG)};
    \node[box, right=2.8cm of graph] (policy) {Eviction Policy \\ \footnotesize(graduated,\\\footnotesize LLM-free)};
    \node[box, below=2.0cm of graph] (tokens) {Token Accounting};

    \draw[arr] (agent) -- node[above, font=\scriptsize] {annotations} (graph);
    \draw[arr] (graph) -- node[above, font=\scriptsize] {candidates} (policy);
    \draw[arr] (policy.west) to[out=210, in=330]
               node[below, font=\scriptsize] {evictions} (graph.east);
    \draw[arr] (graph.south) -- (tokens.north);
    \draw[arr] (tokens.east) -| node[right, font=\scriptsize, pos=0.6] {budget} (policy.south);
  \end{tikzpicture}
  \caption{Components of \cwl. The agent annotates its trajectory as it works; the episode graph accumulates these annotations; the eviction policy consults token accounting and evicts content in priority order when the budget is exceeded.}
  \label{fig:arch}
\end{figure}

\cwl\ has three components: (i) an \emph{annotation protocol} by which the agent marks episode boundaries and dependencies as it works; (ii) an \emph{episode graph} that accumulates these annotations into a typed DAG over the session; and (iii) an \emph{eviction policy} that is invoked whenever token accounting indicates the budget has been exceeded. Figure~\ref{fig:arch} sketches the relationship.

We describe each component in turn.

\subsection{The Annotation Protocol}
\label{sec:arch-annotation}

The agent is given a single tool, \texttt{delimiter}. It produces no output that affects task behavior; its sole purpose is to segment the trajectory. The tool accepts the following schema:

\begin{itemize}
  \item \texttt{action}: \texttt{"start"} $|$ \texttt{"end"} (required).
  \item \texttt{name}: string, required when \texttt{action} is \texttt{"start"}.
  \item \texttt{type}: \texttt{"expl"} $|$ \texttt{"act"}, required when \texttt{action} is \texttt{"start"}.
  \item \texttt{dependencies}: string array, required when starting an \texttt{"act"} chunk; must reference names of earlier \texttt{"expl"} chunks.
  \item \texttt{description}: string, required when ending an \texttt{"expl"} chunk; rejected when ending an \texttt{"act"} chunk.
\end{itemize}

The four canonical calls are therefore:

\begin{enumerate}
  \item \textbf{Start exploration:} \verb|{"action":"start","name":"…","type":"expl"}|
  \item \textbf{Start action:} \verb|{"action":"start","name":"…","type":"act","dependencies":["expl-a",…]}|
  \item \textbf{End exploration:} \verb|{"action":"end","description":"…"}|
  \item \textbf{End action:} \verb|{"action":"end"}|
\end{enumerate}

\paragraph{Episode types.} Two types are distinguished:

\begin{itemize}
  \item \emph{Exploratory} (\texttt{expl}) episodes gather information. They contain tool calls whose outputs inform later decisions but whose raw content is typically not needed once that inference is made: search results, directory listings, file reads performed to orient. When closed, the agent supplies a \texttt{description} summarizing what was learned; this description is the only content retained after full eviction.
  \item \emph{Action} (\texttt{act}) episodes take action. They contain the writes, edits, and tool calls that constitute the agent's actual work. Their effects are persisted in the environment --- a file edit is durable regardless of whether the episode remains in context --- making them the first candidates for eviction.
\end{itemize}

\paragraph{Dependency declarations.} When the agent opens an action episode, it declares the exploratory episodes it depends on via the \texttt{dependencies} field --- a list of names of previously closed \texttt{expl} chunks whose information the action episode is consuming.

This encodes the fact that the action episode's correctness relies on the exploratory one having happened, and it is what allows the eviction policy to preserve causal antecedents (Principle~\ref{prin:causal}). Exploratory episodes do not declare dependencies --- by design, exploration is the frontier and has nothing behind it.

\subsection{The Episode Graph}
\label{sec:arch-graph}

The annotations accumulate into a directed acyclic graph $G = (V, E)$ where vertices $v \in V$ are episodes and an edge $(u, v) \in E$ indicates that action episode $v$ declared a dependency on exploratory episode $u$. The graph is append-only during normal operation; compression removes nodes but never alters edges.

Three invariants are maintained at all times:

\begin{enumerate}
  \item \textbf{Acyclicity.} Dependencies may only reference already-closed episodes, so cycles cannot form.
  \item \textbf{Typed edges.} All edges go from exploratory to action episodes. Action-to-action and exploratory-to-exploratory dependencies are not expressible in the protocol, which keeps the graph's structure simple and the eviction policy tractable.
  \item \textbf{Prologue protection.} Content predating the first \texttt{delimiter} \texttt{start} call --- the system prompt, tool definitions, and any initial user turns --- is treated as a protected prologue and is never part of the graph. It is not eligible for eviction under any circumstances.
\end{enumerate}

A fourth piece of state is the set of \emph{active} episodes: those for which a \texttt{delimiter} \texttt{start} call has been made but the corresponding \texttt{end} call has not yet arrived. Active episodes are never eligible for eviction --- the agent is in the middle of using them.

\subsection{The Eviction Policy}
\label{sec:arch-policy}

When token accounting reports that the current context exceeds the configured threshold, the eviction policy runs. It operates in a loop: each iteration performs the smallest possible eviction that might help, then re-checks the budget. The loop exits as soon as the budget is satisfied or no further evictions are possible.

Within a single iteration, the policy considers all eligible episodes --- closed, non-prologue, with no un-evicted dependents --- and selects the one to operate on according to a simple rule: the oldest eligible action episode, if any exists; otherwise the oldest eligible exploratory episode. The priority for action over exploration reflects what each type contains: exploratory episodes hold the accumulated context information --- search results, file contents, environmental state --- that the agent needs to reason correctly, while action episodes primarily record edits and writes whose effects are already persisted in the environment and can be reconstructed by inspecting the environment.

The dependency constraint is the non-obvious part. An exploratory episode $u$ is only eligible for eviction if every action episode $v$ with $(u, v) \in E$ has itself already been fully evicted. This prevents the situation in which an exploratory episode is dropped while an action episode that depended on it is still in context --- a state in which the agent would have a decision on record but no access to the reasoning that produced it.

Within the selected episode, content is stripped in ordered \emph{levels} of increasing aggressiveness. Each level is attempted, token usage is re-evaluated, and the loop exits if the budget is met. The levels are:

\begin{enumerate}
  \item \textbf{Reasoning trace stripping} (exploratory episodes only). Extended chain-of-thought content is removed first. Reasoning traces are often an insignificant fraction of an exploratory episode's token footprint, and their conclusions are by construction reflected in the tool calls and decisions that follow them.
  \item \textbf{Bulk-output stripping.} Large, enumerable tool outputs --- search results, directory listings (like grep and glob) --- are removed entirely.
  \item \textbf{Intermediate artifact stripping.} Smaller tool interactions --- file reads, bash commands and their outputs --- are removed entirely.
  \item \textbf{Full episode removal.} If stripping is exhausted and the budget is still exceeded, the episode is removed in its entirety.
\end{enumerate}

Throughout, user turns are preserved exactly, as they define the long-horizon trajectory and the ground-truth requirements and instructions from the user (Principle~\ref{prin:user}). Algorithm~\ref{alg:evict} gives the pseudocode.

\begin{algorithm}[H]
\caption{\cwl\ eviction pass}
\label{alg:evict}
\begin{algorithmic}[1]
  \Require \texttt{episodeGraph}, \texttt{tokenBudget}
  \Statex
  \While{\texttt{countTokens(episodeGraph)} $>$ \texttt{tokenBudget}}
    \Statex \quad\textit{\small// Candidates: closed episodes whose dependents are all already evicted}
    \State \texttt{candidates} $\gets$ episodes in \texttt{episodeGraph} where
    \Statex \qquad\qquad \texttt{episode.isClosed}
    \Statex \qquad\qquad \textbf{and} \texttt{not episode.isPrologue}
    \Statex \qquad\qquad \textbf{and} \texttt{not episode.isActive}
    \Statex \qquad\qquad \textbf{and} all dependents of \texttt{episode} are fully evicted
    \If{\texttt{candidates.isEmpty()}}
      \State \textbf{break} \Comment{nothing safe left to evict; budget cannot be met}
    \EndIf
    \Statex \quad\textit{\small// Prefer action episodes (effects already persisted) over exploratory ones}
    \State \texttt{target} $\gets$ oldest \texttt{ACT} episode in \texttt{candidates},
    \Statex \qquad\qquad\quad or oldest \texttt{EXPL} episode if no \texttt{ACT} exists
    \Statex \quad\textit{\small// Try stripping levels from least to most destructive}
    \For{\texttt{level} \textbf{in} [\texttt{STRIP\_REASONING}, \texttt{STRIP\_BULK\_OUTPUT}, \texttt{STRIP\_INTERMEDIATE}, \texttt{REMOVE\_EPISODE}]}
      \If{\texttt{level} is applicable to \texttt{target.type}}
        \State strip \texttt{target} at \texttt{level}: remove content
        \If{\texttt{countTokens(episodeGraph)} $\leq$ \texttt{tokenBudget}}
          \State \textbf{return} \Comment{budget satisfied}
        \EndIf
      \EndIf
    \EndFor
  \EndWhile
\end{algorithmic}
\end{algorithm}

\section{Design Tradeoffs}
\label{sec:tradeoffs}

\cwl\ is not free. Its costs are paid in places different from where compaction pays its costs, and it is worth making the tradeoffs explicit.

\paragraph{Annotation burden.} The agent must call the \texttt{delimiter} tool at every episode boundary. This is a small overhead on each open and close, and a larger one at the start of each action episode, where dependencies must be declared. We accept this cost because the alternative --- inferring structure post-hoc --- is the very thing that makes compaction fragile.

\paragraph{Dependence on annotation quality.} If the agent mis-declares an action episode as exploratory, it will be preserved longer than necessary; if it mis-declares an exploratory episode as action, it will be evicted earlier than it should be, potentially stripping live context the agent still needs. If it omits a dependency, a needed exploratory episode may be evicted while the action episode that relied on it remains. The eviction policy is only as good as the annotations it reads. We mitigate this with prompt-level guidance and with a default that biases toward conservatism (unannotated content is treated as exploratory-with-unknown-dependencies, preserving it until action episodes are exhausted), but the fundamental dependence remains.

\paragraph{Episode graph overhead.} The episode graph is a live data structure maintained in the host process throughout the session. For sessions with hundreds of episodes or deep dependency chains, graph traversal during eviction passes is non-negligible. In practice we expect sessions to contain tens of episodes at most, keeping traversal cheap; but workloads with unusually fine-grained annotation --- many short episodes --- may accumulate graph state that warrants periodic compaction of fully-evicted subgraphs.

\paragraph{KV cache invalidation.} Modern LLM inference APIs exploit \emph{prefix caching} (KV caching): if the token prefix of a new request matches a previously cached one, the key-value activations from that prefix are reused, substantially reducing the compute cost of a request. \cwl's eviction policy modifies the context in place, which changes the prompt prefix presented to the inference engine. Every eviction therefore invalidates the cached KV state for the affected prefix and all content that follows it.

At low eviction frequency this is acceptable: cache misses are the normal fallback and the net cost is no worse than an uncached request. The problem arises under sustained token pressure. When the session operates continuously near the budget threshold, evictions occur on every turn or every few turns. The inference system then enters a regime in which it consistently pays the \emph{cache-write} cost --- incurred on every request whose prefix is new --- without ever amortizing it through \emph{cache-read} savings, because each new eviction invalidates the entry before the next request can reuse it.

In the worst case, \cwl\ at maximum utilization is net-negative for caching: more is spent on cache writes that will be immediately discarded than would have been spent on uncached inference. This is, in a narrow sense, worse than compaction, which --- despite its other flaws --- produces a single stable prefix after each compaction pass and can sustain a cache hit rate until the next pass. \cwl's graduated, incremental eviction fragments the prefix more frequently.

In our benchmarking we identified a potential solution, though not without a further tradeoff. We denote the token budget ceiling as $\tau$ and capped it at 80{,}000 tokens --- approximately 30\% of the context window of the models under evaluation. Because eviction keeps the active token count stable near this ceiling rather than allowing it to grow toward the full window, the prefix seen by the inference engine is effectively constant across turns: new content is added at the tail while an equal volume is evicted at the head, and the stable bulk of the prefix accumulates cache hits. The result was a 20--70\% reduction in inference cost relative to uncapped sessions, with the range reflecting differences in task type and session nature: tasks with repetitive, structurally similar turns (e.g., iterative code editing) benefited most, while sessions with highly variable tool outputs benefited least. Setting $\tau$ is a dial with three dimensions: \emph{cost} (lower $\tau$ reduces per-turn inference cost by stabilising the prefix), \emph{look-back capability} (lower $\tau$ evicts content sooner, requiring re-exploration for tasks with long natural look-back windows), and \emph{model quality} (lower $\tau$ keeps prompts in the regime where attention and hallucination rates are favourable). The first and third dimensions both favour a lower $\tau$; only the second pulls against them. This means the tradeoff is less severe than a pure cost-versus-capability framing suggests: a tighter budget is simultaneously cheaper and qualitatively better, up to the point where necessary look-back is impaired. The optimal $\tau$ is workload-dependent, but the quality dimension argues for erring toward a lower value rather than a higher one.

\section{Empirical Evaluation}
\label{sec:eval}

\subsection{Setup}

We implemented \cwl\ as an extension to a fork of the open-source agent harness \texttt{pi.dev}; the implementation is publicly available at \url{https://github.com/Kiz8-Team/pi-cwl}~\citep{kiz8team2025picwl}. The fork adds the \texttt{delimiter} tool to the agent's tool set, maintains the episode graph in the harness process, and runs the eviction policy on every turn after token accounting. All experiments use GPT-5.4 as the underlying model.

Beyond the \cwl\ integration, we made several harness-level optimizations that meaningfully reduce baseline token consumption and warrant description in their own right:

\begin{itemize}
  \item \textbf{Minimal system prompt.} The system prompt was stripped to a bare minimum, occupying under 1{,}000 tokens. Contemporary agent harnesses are substantially heavier: Claude Code's initial prompt consumes approximately 20{,}000 tokens, and similar overhead is present in other widely used harnesses. Eliminating this overhead directly expands the effective budget available for task context.

  \item \textbf{Git-status injection.} The current \texttt{git status} is injected into the prompt at each turn. This gives the agent a reliable, low-cost snapshot of repository state without requiring a shell call, reducing redundant environment-probing tool calls.

  \item \textbf{Glob and grep with gitignore awareness.} Pattern-based search tools (\texttt{glob}, \texttt{grep}) were instrumented to respect \texttt{.gitignore}, and explicit usage instructions were added to guide the agent toward targeted pattern searches rather than broad file reads. Harnesses that lack this --- notably OpenAI's Codex, which does not expose pattern search and relies on per-line file reads --- incur substantially higher token costs for equivalent codebase orientation. In our measurements, these search optimizations alone produced a 3--6$\times$ reduction in token consumption relative to Codex-style harnesses on equivalent tasks.
\end{itemize}

These optimizations had a secondary effect relevant to \cwl\ evaluation: the combined token efficiency made it difficult for the agent to overflow the 80{,}000-token budget on shorter benchmarks, which informed the design of the evaluation protocol below.

\subsection{Evaluation Methodology}

We evaluated \cwl\ across four benchmarks: \textbf{Terminal Bench 2.0}~\citep{merrill2026terminalbench}, a suite of 89 agentic tasks covering terminal-based coding and system interaction; \textbf{SWE Bench Lite}~\citep{jimenez2024swebench}, evaluated on a randomly sampled subset of 50 tasks from the full 300-task set; \textbf{Recovery Bench}~\citep{recoverybench2025}, which tests the agent's ability to recover from failure states mid-task; and \textbf{LongCLI Bench}~\citep{feng2026longcli}, which covers long-horizon command-line interaction tasks.

\paragraph{Protocol.} The standard evaluation protocol for all four benchmarks runs each task in a separate, fresh agent session. This isolates tasks from one another but cannot exercise context management --- no session accumulates enough history to approach any reasonable token budget.

We applied a uniformly harder protocol for the \cwl\ condition across every benchmark: \emph{all tasks are executed sequentially in a single, uninterrupted agent session} with \cwl\ active throughout. The session begins with the first task; the agent completes it and moves directly to the next without any session reset; the episode graph accumulates the full cross-task trajectory. The baseline condition always uses the standard per-task isolated-session protocol --- a fresh context for each task, matching the protocol used by all published leaderboard entries.

This asymmetry is the central point of evaluation. The baseline never faces context pressure. The \cwl\ condition accumulates a full benchmark-length trajectory under a fixed 80{,}000-token budget. Comparisons to leaderboard scores are therefore not apples-to-apples: the question under evaluation is not how \cwl\ ranks, but whether a single accumulating session can sustain task performance across an entire benchmark suite.

\subsection{Results}

\begin{figure}[t]
  \centering
  \includegraphics[width=0.85\textwidth]{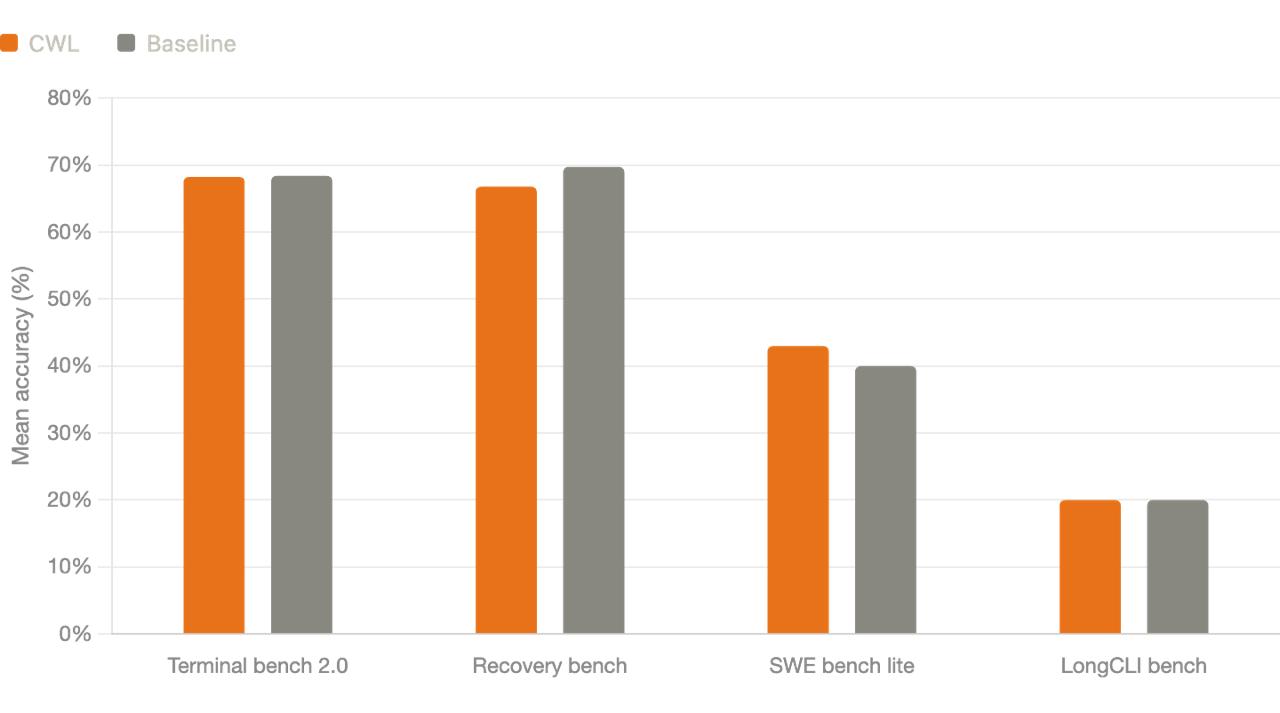}
  \caption{Mean accuracy (\%) for \cwl\ (single session, 80k token budget) and the individual-sessions baseline across four benchmarks. Terminal Bench 2.0: $n=5$; SWE Bench Lite, Recovery Bench, LongCLI Bench: $n=3$.}
  \label{fig:accuracy}
\end{figure}

We report mean accuracy across independent runs: $n=5$ for Terminal Bench 2.0, $n=3$ for all other benchmarks. Figure~\ref{fig:accuracy} summarizes results across all four benchmarks.

Results by benchmark: Terminal Bench 2.0 --- \cwl\ \textbf{68.25\%}, baseline \textbf{68.40\%}; SWE Bench Lite (50-task sample) --- \cwl\ \textbf{43.00\%}, baseline \textbf{40.00\%}; Recovery Bench --- \cwl\ \textbf{66.80\%}, baseline \textbf{69.75\%}; LongCLI Bench --- \cwl\ \textbf{20.00\%}, baseline \textbf{20.00\%}.

Across all four benchmarks, the \cwl\ and baseline conditions differ by at most 3 percentage points in either direction. These margins are within run-to-run variance and should not be interpreted as directional signal. The principal result is consistent across all benchmarks: \cwl\ produces \emph{no measurable degradation} in task accuracy relative to the individual-session baseline, despite operating under a substantially harder regime.

\paragraph{Token consumption and cost.} Across all benchmarks, \cwl\ and the baseline showed little to no difference in total token consumption or inference cost --- the shorter benchmark suites did not generate enough sustained context pressure to produce a measurable gap. The exception is Terminal Bench 2.0, where the 89-task single-session run accumulated sufficient scale to make the effect visible: \cwl\ processed \textbf{over 80 million tokens} across the full sequence at a total inference cost of approximately \textbf{\$55 per complete run}. Maintaining a stable active token count near the budget ceiling --- new content entering at the tail while eviction keeps the prefix bounded --- reduced per-turn inference cost relative to an uncapped session. Observed cost reduction ranged from 20--70\% relative to uncapped sessions, depending on task structure; the mechanism and the range are discussed in detail in Section~\ref{sec:tradeoffs}.

\subsection{Budget Sensitivity}

We varied $\tau$ to characterize the cost-capability tradeoff described in Section~\ref{sec:tradeoffs}. The principal findings are:

\begin{itemize}
  \item \textbf{Budgets above 120{,}000 tokens} produced a sharp increase in inference cost with no corresponding accuracy improvement. The larger context creates more prefix instability per eviction pass, and the benefit of increased look-back does not offset the cache-miss penalty.

  \item \textbf{Budgets around 50{,}000 tokens} reduced inference costs by up to $3\times$ relative to the 120k+ baseline, with no measurable degradation in task accuracy. However, wall-clock time per task increased by up to $2\times$. The mechanism is visible in the trace: the aggressive budget forced eviction of still-relevant exploratory content, and the agent subsequently re-explored the same codebase regions it had already examined. Accuracy was maintained because re-exploration recovered the evicted information, but the extra tool calls added latency and token cost for the re-exploration itself.
\end{itemize}

These results suggest that $\tau \in [80{,}000,\,120{,}000]$ tokens is near a Pareto frontier for the tasks under evaluation: reducing it recovers cost at the price of time, while increasing it adds cost without adding capability. The optimal $\tau$ will differ for workloads with longer natural look-back windows or less code-exploration structure.

\subsection{Case Study: Cross-Task Dependency on Real-World Repositories}
\label{sec:eval2}

Terminal Bench 2.0 tasks are independent: even in our sequential protocol, each task is self-contained and the agent can succeed on task $k$ without remembering anything about task $k-1$. To probe \cwl\ under genuine cross-task dependency --- where later tasks require the agent to build on artifacts and decisions made in earlier ones --- we ran a targeted experiment on real-world open-source repositories. This is a qualitative case study rather than a controlled benchmark; the goal is to illustrate concretely how structured eviction differs from compaction when causal context must be preserved across task boundaries.

\paragraph{Setup.} We selected three repositories of increasing context complexity: \textbf{Excalidraw} (a TypeScript web application), \textbf{Redis} (a C systems codebase), and the \textbf{Linux kernel} (a large C codebase with a custom build and test pipeline). For each repository we designed a sequence of 3--4 tasks that either build directly on one another or share enough structural context that a model working across them benefits from retaining trajectory. The task sequences were:

\begin{itemize}
  \item \textbf{Excalidraw} (3 tasks): implement a color wheel picker; implement a callout shape; implement a pages system (multiple independent canvases).
  \item \textbf{Redis} (3 tasks): add \texttt{NOVALUES} flag support to \texttt{HGETALL}; extend the memory profiler to cover the new flag; write a test suite covering both.
  \item \textbf{Linux kernel} (4 tasks): implement a PID summary syscall; expose its output as a materialized \texttt{/proc} entry for each PID; build the kernel; set up QEMU and validate both prior tasks end-to-end.
\end{itemize}

We ran each repository's task sequence twice: once with \cwl\ active ($\tau = 80{,}000$ tokens, single session, GPT-5.4), and once with the same harness and model but with the harness's default compaction in place of \cwl. The context-management strategy is the only thing that differs between the two runs.

\paragraph{Token efficiency.} \cwl\ was substantially more token-efficient than the compaction run. Inference cost was \textbf{23\% lower} with \cwl\ despite both completing the same task sequences. The mechanism is the same as in the benchmark evaluation above: \cwl\ holds active tokens near the 80k ceiling and evicts completed action episodes, while compaction allows the context to grow until a threshold fires and replaces the history with a summary --- a large model call that itself consumes tokens and resets the KV cache state.

\paragraph{Observations.} Outcomes by repository were as follows:

\begin{itemize}
  \item \textbf{Excalidraw.} Both runs completed all three tasks correctly. The tasks are loosely related and the codebase is well-structured, making cross-task dependency shallow; both context strategies were sufficient.

  \item \textbf{Redis.} Both runs completed all three tasks correctly. The \texttt{NOVALUES} extension and the test suite depend on the same data-structure decision made in the first task; both strategies retained enough context to apply it consistently.

  \item \textbf{Linux kernel.} Both runs completed tasks 1--3 (PID summary syscall, materialized \texttt{/proc} entry, and kernel build). Both encountered difficulty in task 4 (QEMU boot and validation). \cwl\ became stuck in a loop while attempting to configure the emulator; manual assistance resolved the impasse and the agent completed the validation run. The compaction run also struggled with QEMU configuration, as compaction had discarded build and environment context accumulated in prior tasks, complicating recovery.

  The most informative difference appeared in task 2. \cwl\ retained the full episode graph through task 1 and implemented the \texttt{/proc} entry with direct access to the kernel data structures, helper functions, and registration patterns from the syscall implementation. The compaction run's summarization pass, which fired after task 1, retained a prose description of what the syscall did but discarded the structural detail needed to wire the \texttt{/proc} entry to it correctly. The compaction run completed the connection only after additional exploratory tool calls to reconstruct context that \cwl\ had preserved intact.
\end{itemize}

\paragraph{Takeaway.} The Linux kernel case concretely illustrates the cost of compaction's structural information loss. When the baseline's compaction pass fired after task 1, it discarded the causal structure linking the PID syscall implementation to the \texttt{/proc} entry --- the failure mode identified in Section~\ref{sec:intro}. The baseline recovered through re-exploration, but at the cost of redundant tool calls and additional latency. \cwl's episode graph preserved the structural detail intact and delivered it to task 2 without additional cost. The 23\% inference cost reduction, combined with avoided re-exploration overhead, reflects this: structured eviction spends less on compression and less on recovering what compression discarded.

\section{Limitations and Open Questions}
\label{sec:limitations}

We note several open questions that the current design does not resolve.

\emph{Dependency granularity.} The protocol allows action episodes to depend on whole exploratory episodes. It does not allow dependencies on specific tool calls within an episode. This is a deliberate simplification; finer granularity would complicate both the annotation interface and the eviction policy. Whether the coarser granularity is sufficient in practice is an empirical question we are evaluating.

\emph{Non-linear trajectories.} Some agents branch, backtrack, or operate over multiple parallel subtasks. The current design assumes a single linear stream of episodes and expresses non-linearity only through the dependency edges. Whether richer structure (e.g., subgraphs, subtask roots) is needed is open.

\emph{Effect on model reasoning behavior.} Introducing the \texttt{delimiter} tool and its associated instructions may alter how a model reasons about and structures its exploration, independently of the eviction policy itself. In our evaluations we observed a range of behavioral changes that we were unable to conclusively attribute to \cwl: in some sessions the model rushed through exploration with less thoroughness than it exhibited without \cwl; in longer sessions it occasionally over-explored, revisiting already-annotated material or looping over actions without apparent progress. Because these behaviors overlap with ordinary model hallucination and planning failures, we could not isolate \cwl\ as their cause in individual cases. We tentatively attribute them to mild confusion introduced by the annotation protocol --- an additional layer of meta-reasoning the model must perform alongside the task itself. If this attribution is correct, the effect is likely to diminish as underlying model capability improves; more capable models should be better able to treat the annotation protocol as a lightweight bookkeeping obligation rather than a reasoning burden.

\section{Conclusion}
\label{sec:conclusion}

We have presented \cwl, a context-management scheme that gives long-horizon LLM agents an effectively unbounded working horizon. Built on three ideas --- the agent annotates its trajectory as it works, the annotations form a typed dependency graph, and a deterministic graduated eviction policy walks that graph when the token budget is exceeded --- \cwl\ avoids four well-known limitations of summarization-based compaction (unpredictable lossiness, structural destruction, blocking cost, and compression-induced hallucination) by construction, and generalizes recency truncation by making eviction semantically aware of the dependency graph. Each architectural choice traces to one of six design principles; empirical evaluation demonstrates capability parity with per-task isolated sessions over an 89-task, 80-million-token single agent session.

\bibliographystyle{plainnat}
\bibliography{references}

\end{document}